\documentclass[runningheads,a4paper]{llncs}
\usepackage{amssymb}
\usepackage{graphicx}
\usepackage{amsmath}
\usepackage{algorithm}
\usepackage{algorithmic}
\DeclareMathOperator*{\argmax}{argmax}
\DeclareMathOperator*{\argmin}{argmin}
\usepackage[lined,algonl,boxed,algo2e]{algorithm2e}
\usepackage{caption}
\usepackage{subcaption}
\usepackage{url}
\usepackage[colorlinks]{hyperref}
\usepackage{lipsum}

\usepackage{xcolor}
\hypersetup{colorlinks, linkcolor={red!50!black}, citecolor={red!50!black},  urlcolor={red!50!black}}
\usepackage[switch*]{lineno}
   
\newcommand{\keywords}[1]{\par\addvspace\baselineskip
\noindent\keywordname\enspace\ignorespaces#1}
\usepackage{amsmath}
\usepackage{amssymb}
\urldef{\mailsa}\path|{aybuke.ozturk, stephane.lallich, jerome.darmont}@univ-lyon2.fr|
\urldef{\mailsb}\path|yona.waksman@mom.fr|

\begin{document}

\mainmatter  

\title{MaxMin Linear Initialization \\ for Fuzzy C-Means}
\author{Ayb\"uke \"Ozt\"urk$^{*, **}$
\and  St\'ephane Lallich$^{*}$ \\
J\'er\^{o}me Darmont$^{*}$ 
\and Sylvie Yona Waksman$^{**}$}
\institute{Universit\'e de Lyon, Lyon 2, France\\
$^{*}$ERIC EA 3083 (5 avenue Pierre Mend\`es France, F69676 Bron Cedex) \\
$^{**}$ArAr UMR 5138 (7 rue Raulin, 69365 Lyon Cedex 7)\\ \mailsa\\
\mailsb}

\maketitle

\begin{abstract}

Clustering is an extensive research area in data science. The aim of clustering is to discover groups and to identify interesting patterns in datasets. Crisp (hard) clustering considers that each data point belongs to one and only one cluster. However, it is inadequate as some data points may belong to several clusters, as is the case in text categorization. Thus, we need more flexible clustering. Fuzzy clustering methods, where each data point can belong to several clusters, are an interesting alternative. Yet, seeding iterative fuzzy algorithms to achieve high quality clustering is an issue. In this paper, we propose a new linear and efficient initialization algorithm \textit{MaxMin Linear} to deal with this problem. Then, we validate our theoretical results through extensive experiments on a variety of numerical real-world and artificial datasets. We also test several validity indices, including a new validity index that we propose, \textit{Transformed Standardized Fuzzy Difference} (TSFD). 

\keywords{\textit{Clustering, Fuzzy C-Means, Seeding, Initialization, Maxmin Linear Method, Validity Indices}}
\end{abstract}

\section{Introduction}

Clustering is a useful technique for grouping a set of unlabelled data points (instances) described by attributes (variables), such that points belonging to the same cluster (group) have similar characteristics, while points in different clusters have dissimilar characteristics. There are several types of clustering schemes, such as crisp, overlapping or fuzzy partitions, and hierarchies. Crisp clustering considers that each data point belongs to one and only one cluster. Contrary to crisp clustering, fuzzy clustering \cite{ruspini1970numerical} considers that a data point can belong to more than one cluster. There are some situations where fuzzy clustering is very useful. For instance, let us consider three clusters achieved when categorizing textual documents: an economy cluster (topic), an energy cluster, and a politics cluster. Then a document containing the keyword ``petrol" could belong to all three clusters. Moreover, fuzzy clustering helps opening a discussion with domain experts regarding clustering results. 

The primary objective of our paper is to avoid using highly complex clustering methods. One solution is to use iterative fuzzy methods such as Fuzzy C-Means (FCM) and Fuzzy K-Medoids. Both methods adapt the principle of the K-Means algorithm \cite{macqueen1967some}. FCM, proposed by \cite{dunn1973fuzzy} and extended by \cite{bezdek1984fcm}, applies on numerical data, while Fuzzy K-Medoids  \cite{kaufman1990partitioning} applies on categorical data. Since numerical data are the most common case, we choose to experiment our proposals with FCM.

The aim of the FCM algorithm is to minimize the fuzzy within-inertia $FW$ (see Equation~\ref{eqFW}). Fuzzy inertia $FI$ (see Equation~\ref{eqFI}) composes of the $FW$ and the fuzzy between-inertia $FB$ (see Equation~\ref{eqFB}). $FW$, $FI$, and $FB$ are computed from a membership matrix $U$, which stores the  membership coefficients $u_{ik}$ of data point $i$ to cluster $k$. Note that $FI$ = $FW$ + $FB$. Moreover, $FI$ is not constant because it depends on $u_{ik}$ value. When $FW$ changes, the values of $FI$ and $FB$ also change.

\begin{equation} \label{eqFW}
FW = \sum_{i=1}^n \sum_{k=1}^K  u_{ik}^m d^2 (x_i,c_k)
\end{equation}

\begin{equation} \label{eqFI}
FI = \sum_{i=1}^n \sum_{k=1}^K  u_{ik}^m d^2 (x_i,\overline{x})
\end{equation}

\begin{equation} \label{eqFB}
FB = \sum_{i=1}^n \sum_{k=1}^K  u_{ik}^m d^2 (c_k,\overline{x})
\end{equation}

where $n$ is the number of instances, $K$ is the number of clusters, $m$ is the fuzziness coefficient (by default, $m = 2$. If $m = 1$, clustering is crisp. If $m > 1$, clustering becomes fuzzy), $c_k$ is the center of the $k^{th}$ cluster $\forall$ k, $1 \leq k \leq K$, $\overline{x}$ is the grand mean (the arithmetic mean of all data, see Equation~\ref{Equation_arithmeticMean}), and function $d^2()$ computes the squared Euclidean distance.

\begin{equation} \label{Equation_arithmeticMean}
\overline{x} = \frac{1}{n}\sum_{i=1}^n x_i
\end{equation}

FCM starts by choosing $K$ data points as initial centroids of the clusters. Then, membership matrix values $u_{ik}$ (see Equation~\ref{FCMMembershipM_eqn}) are assigned to each data point in the dataset. Centroids of clusters $c_k$ are updated based on Equation~\ref{FCMCentroid_eqn} until a termination criterion is reached successfully. In FCM, this criterion can be a fixed number of iterations $t$, e.g., $t = 100$. Alternatively, a threshold $\epsilon$ can be used, e.g., $\epsilon = 0.0001$. Then, the algorithm stops when the relative difference of objective function $< \epsilon$.

\begin{equation} \label{FCMMembershipM_eqn}
u_{ik} = \frac{1}{\sum_{j=1}^K (\frac{\|x_i-c_k\|^2}{\|x_i-c_j\|^2})^\frac{1}{m-1}} 
\end{equation} 

\begin{equation} \label{FCMCentroid_eqn}
c_k = \frac{\sum_{i=1}^n(u_{ik}^m)x_i}{\sum_{i=1}^n(u_{ik}^m)}
\end{equation}

When using FCM, an important point is the way of choosing $K$ data points as initial centroids (seeds). An efficient initialization method should be linear, so that the FCM algorithm stays linear, too. Then, the initialization method must be evaluated using validity indices that are well suited to the fuzzy case.

To obtain a good validated clustering result, one has to minimize intra-cluster distance (compactness) and at the same times, one has to maximize inter-cluster distance (separability). The more often, proposed clustering validity indices associate a compactness index with a separability index. 

Thence, we propose in this paper (1) a linear and efficient initialization method for FCM clustering called \textit{MaxMin Linear}. Moreover, to compare our proposal with several initialization methods from the literature, we also propose (2) a new clustering validity index called \textit{Transformed Standardized Fuzzy Difference} (TSFD), which is tailored to the fuzzy case. We perform validation experiments on several numerical real-world and artificial datasets. 

The remainder of this paper is organized as follows. Section~\ref{sec:RelatedWorks} presents initialization methods for iterative clustering and several clustering validity methods proposed in the literature. Sections~\ref{sec:ContributionMaxMinLinear} and \ref{sec:ContributionValidityIndex} detail our contributions, i.e., the \textit{MaxMin Linear} initialization method and the TSFD validity index, respectively. Section~\ref{sec:Experiments} deals with the experimental evaluation of the \textit{MaxMin Linear} initialization method on several datasets, using several validity indices, including TSFD. Finally, we conclude this paper and provide some perspectives in Section~\ref{sec:Conclusion}.

\section{Related Works}
\label{sec:RelatedWorks}

Most initialization methods are studied through K-Means clustering \cite{macqueen1967some} concepts. We have reviewed various works from the literature, including much-cited papers \cite{steinley2007initializing,maitra2011systematic,celebi2013comparative}. In our study, we make use of commonly mentioned linear methods from these three papers. 

The first initialization method by \cite{macqueen1967some} uses the first $K$ data points as centroids. This method is sensitive to the order of data. It is used by default in SPSS \cite{noruvsis2012ibm}. The second method by MacQueen (\textit{MacQueen2}) takes $K$ random data points as centroids. Moreover, \cite{faber1994clustering} proposes to perform multiple relaunches of \textit{MacQueen2}. Among the different relaunches, the one that optimizes $FW$ (Equation~\ref{eqFW}) is considered the best candidate. This method is the standard way for initializing clusters. Its main disadvantage is that already selected points are not considered when a new seed is chosen. The second disadvantage is that outliers can be chosen. On the other hand, multiple runs ensure to improve the quality of the chosen sample. 

Hand et al. \cite{hand2005optimising}, propose an extension of Faber's method that starts with a random set of seeds. It suggests iteratively modifying the partition by randomly moving some points to other clusters. The partition  minimizing $FW$ is chosen as the best candidate. To move each data point to another random cluster, a probability $\alpha$, e.g., $\alpha = 0.3$, must be set. The method is only interesting if parameter $\alpha$ is fixed for different datasets.

Bradley and Fayyad's method \cite{bradley1998refining} starts by randomly partitioning the dataset into $J$ subsets. Then, each subset is clustered with the K-Means algorithm using \textit{MacQueen2} initialization. \textit{MacQueen2} produces $J$ sets of centers, each containing $K$ points. The centers of clusters are combined into a superset. Then, the superset is clustered by K-Means $J$ times. Each time, K-Means is initialized with a different center set, and members of the center set that give the smallest $FW$ are selected as final centers. 

The PCA-Part method \cite{su2007search} uses a divisive hierarchical approach based on Principal Component Analysis (PCA) \cite{wold1987principal}. The method starts with a single cluster containing the whole dataset. Then, it iteratively divides clusters with respect to $FW$. Clusters are divided into two sub-clusters by using a hyperplane that is orthogonal to the principal eigenvector of the cluster covariance matrix. The division process ends after $K$ clusters are obtained.  

The K-Means++ method \cite{arthur2007k} selects centroids based on a distance probability to the nearest center. First, it randomly selects an initial center $c_1 = x$ from the data point set $X$. Then, $d(x)$ is denoted as the shortest euclidean distance from $x$ to its closest center. The next center $c_i$ is randomly selected as $c_i = x' \in X$ with probability $d(x')^2/\sum d(x)^2$. 

Finally, in the literature, there are other methods having quadratic complexity \cite{lance1967general,astrahan1970speech}. Among quadratic methods, \textit{MaxMin} (also called \textit{Maximin}) \cite{gonzalez1985clustering} is particularly interesting. \textit{MaxMin} first calculates all the paired distances between data points. Then, it chooses two centroids from the data points, which have the greatest distance to each other. Finally, the next centroid is the data point that is the farthest from its centroid. This approach helps decrease $FW$, which improves the homogeneity of clusters. 

To summarize, Hand and Krzanowski \cite{hand2005optimising} rely on user-defined parameters that may not be easy to set. \textit{MacQueen2}, though easy to understand and implement, uses only one random sample. Faber improves the \textit{MacQueen2}'s random sample through relaunches. In K-Means++, the random choice is replaced by a probabilistic choice and cluster homogeneity is taken into account. However, since the probabilistic selection does not always select sufficiently the large enough distance, several probabilistic samples are required and the best centers are selected from all relaunches. 

In contrast, \textit{MaxMin} constructs only one sample by decreasing $FW$ and is thus deterministic. Thus, we can be sure that a chosen center is the best. Yet, it can be less effective than K-Means++ in the presence of outliers. \\

To evaluate initialization methods, we need to use fuzzy validity indices. According to  \cite{wang2007fuzzy}, there are two groups of validity indices. The first group is only based  on membership values and includes the partition coefficient index $V_{PC}$  \cite{bezdek1973cluster} (see Equation~\ref{PC_eqn}; $\frac{1}{K} \leq V_{PC} \leq 1$; to be maximized) and the Chen and Linkens index $V_{CL}$ \cite{chen2001rule} (see Equation~\ref{ChenLinked_eqn}; $0 \leq V_{CL} \leq 1$; to be maximized). 

\begin{equation} \label{PC_eqn}
V_{PC}= \frac{1}{n} \sum_{i=1}^n \sum_{k=1}^K u_{ik}^2
\end{equation} 

\begin{equation} \label{ChenLinked_eqn}
V_{CL} = \frac{1}{n} \sum_{i=1}^n max_{k}(u_{ik}) - \frac{1}{c} 
\sum_{k=1}^{K-1} \sum_{j=k+1}^K \Bigg[\frac{1}{n} \sum_{i=1}^n min(u_{ik}, u_{ij})\Bigg],
\end{equation}

where $c = \sum_{k=1}^{K-1}k $. \\

$V_{CL}$ takes in consideration both compactness (first term of $V_{CL}$) and separability (second term of $V_{CL}$). The second group of fuzzy validity indices is based on associating membership values to cluster centers and data. It includes the adaptation of the Ratio index $V_{FRatio}$ to fuzzy clustering \cite{calinski1974dendrite} (see Equation~\ref{Ratio_eqn}; $0 \leq V_{FRatio} \leq +\infty$; to be maximized), the penalized version of $V_{FRatio}$ index which is the Calinski and Harabasz index $V_{FCH}$ \cite{calinski1974dendrite} (see Equation~\ref{CH_eqn}; $0 \leq V_{FCH} \leq +\infty$; to be maximized), the Fukuyama and Sugeno index $V_{FS}$ \cite{fukuyama1989new} (see Equation~\ref{Fukuyama_eqn}; $-FI \leq V_{FS} \leq FI$; to be minimized), and the Xie and Beni index $V_{XB}$ \cite{xie1991validity,pal1995cluster} (see Equation~\ref{XieBeni_eqn}; $0 \leq V_{XB} \leq FI/n*min\|x_j - v_k\|^2 $; to be minimized).
 
\begin{equation} \label{Ratio_eqn}
V_{FRatio} = FB/FW
\end{equation}

\begin{equation} \label{CH_eqn}
V_{FCH} = \frac{FB/(K-1)} {FW/(n-K)} = \frac {n-K}{K-1} \frac {FB}{FW}
\end{equation}

\begin{equation} \label{Fukuyama_eqn}
V_{FS} = FW - FB
\end{equation}

\begin{equation} \label{XieBeni_eqn}
V_{XB} = \frac{\sum_{k=1}^K \sum_{i=1}^n u_{ik}^m \|x_i - v_k\|^2}{n* min_{j,k} \|v_j - v_k\|^2}
\end{equation}

Among all the above stated validity indices, there is no single validity index that gives the best result for any dataset. Thus, there is room for a new validity index that is specifically tailored for fuzzy validation. This is why we propose the \textit{Transformed Standardized Fuzzy Difference} index.

\section{MaxMin Linear Fuzzy Clustering Initialization Method}
\label{sec:ContributionMaxMinLinear}

\textit{MaxMin}'s simplicity and ability to build homogeneous clusters sounds very interesting. Yet, considering all paired distance between data points makes the method quadratic with respect to the number of data points. Thus, we present in this section an enhancement of \textit{MaxMin} that makes it linear. Before introducing our changes, we first detail how \textit{MaxMin} works in Algorithm~\ref{alg:maxmin} (see Section~\ref{sec:RelatedWorks} for \textit{MaxMin}'s principle).

\begin{algorithm}
\caption{\textit{MaxMin}}
\label{alg:maxmin}
\begin{algorithmic}
	\REQUIRE Set of data points $X = \{x_1,...,x_n\}$
    \REQUIRE Number of clusters $K$
	\STATE \COMMENT{Select the first two centroids $c_1$ and $c_2$}
	\STATE $c_1, c_2 \leftarrow \argmax(d^2(x_i, x_j))$ $i, j = 1,...,n$
	\STATE $K^* \leftarrow 2$ \COMMENT{Number of seeds}
	\STATE \COMMENT{Find the remaining seeds}
	\WHILE{$K^* < K$}
		\FORALL{$x_i \neq c_{k^*}$ $i = 1,...,n$, $k^* = 1,...,K^*$}
			\STATE $d^2_m(x_i) \leftarrow \min(d^2(x_i, c_{k^*}))$
		\ENDFOR
		\STATE $K^* \leftarrow K^* + 1$		
		\STATE $c_{K^*} \leftarrow \argmax(d^2_m(x_i))$ $i = 1,...,n$
	\ENDWHILE
	\RETURN $\{c_{k^*}\}$ $k^* = 1,...,K^*$
\end{algorithmic}
\end{algorithm} 

In \textit{MaxMin Linear}, we first calculate grand mean $\overline{x}$ (see Equation~\ref{Equation_arithmeticMean}). Then, we choose as first centroid the data point that is nearest to $\overline{x}$. The second centroid is the data point that has the largest distance to the first centroid. Thus, complexity remains linear with respect to the number of data points. Afterwards, the choice of the remaining centroids remains the same as in \textit{MaxMin}. \textit{MaxMin Linear} is formalized in Algorithm~\ref{alg:maxminlinear}.

\begin{algorithm}
\caption{\textit{MaxMin Linear}}
\label{alg:maxminlinear}
\begin{algorithmic}
	\REQUIRE Set of data points $X = \{x_1,...,x_n\}$
    \REQUIRE Number of clusters $K$
	\STATE \COMMENT{Select the first two centroids $c_1$ and $c_2$}
    \STATE $\overline{x} \leftarrow \frac{1}{n}\sum_{i=1}^n x_i$ 
    \FOR{$i \leftarrow 1$ \TO $n$}
	    \STATE $d^2_m(x_i) \leftarrow \min(d^2(\overline{x}, x_i))$ 
	\ENDFOR
	\STATE $c_1 \leftarrow \argmin(d^2_m(x_i))$ $i = 1,...,n$
    \FOR{$i \leftarrow 1$ \TO $n$}
	    \STATE $d^2_m(x_i) \leftarrow \max(d^2(c_1, x_i))$ 
	\ENDFOR
	\STATE $c_2 \leftarrow \argmax(d^2_m(x_i))$ $i = 1,...,n$
	\STATE $K^* \leftarrow 2$ \COMMENT{Number of seeds}
	\STATE \COMMENT{Find the remaining seeds}
	\WHILE{$K^* < K$}
		\FORALL{$x_i \neq c_{k^*}$ $i = 1,...,n, k^* = 1,...,K^*$}
			\STATE $d^2_m(x_i) \leftarrow \min(d^2(x_i, c_{k^*}))$
		\ENDFOR
		\STATE $K^* \leftarrow K^* + 1$		
		\STATE $c_{K^*} \leftarrow \argmax(d^2_m(x_i))$ $i = 1,...,n$
	\ENDWHILE
	\RETURN $\{c_{k^*}\}$ $k^* = 1,...,K^*$
\end{algorithmic}
\end{algorithm}

As a final note, the use of \textit{MaxMin Linear} is not limited to use with FCM on numerical data, but also with Fuzzy K-Medoids \cite{park2009simple} for categorical data clustering.
Thus, \textit{MaxMin Linear} can also be applied with heterogeneous data to construct fuzzy clustering ensemble. This makes of \textit{MaxMin Linear} a simple but noteworthy contribution, in our opinion. 

\section{Transformed Standardized Fuzzy Difference Validity Index} 
\label{sec:ContributionValidityIndex}

Several problems must be cleaned up to obtain a good clustering, including evaluation of the validity of the clusters and choosing the number of clusters. However, it is not an easy process. Compactness and separation level might raise problems. Firstly, if the chosen number of clusters is larger than optimal one, some clusters are broken while they could be more compact. Secondly, if the chosen number of clusters is smaller than optimal one, some clusters are merged and while they could be more separated. When it comes to addressing resolve those problems, many cluster validity indices are proposed for fuzzy clustering algorithms. The objective is to find the optimal number of clusters that can validate the best description of the data structure. 

The optimal number of the cluster can be determined by considering the variation of clustering validity index. It is distinguished into two cases: The first case, if the index is not monotonic with the number of clusters, we choose the value of the number of clusters which optimizes the index. The second case, if the index is monotonic, one can prefer to use a penalized version of the index.

In building TSFD, we first consider the difference $FB - FW$, which is similar to FS except for the sign). Unfortunately, $FI = FB + FW$ is not constant and $FB - FW \in [-FI, +FI]$. To take this particularity of fuzzy clustering into account, we propose to standardize $FB - FW$ by considering \textit{Standardized Fuzzy Difference} $SFD = (FB - FW) \div FI$ instead. $SFD \in [-1, +1]$.

Finally, to obtain an index belonging to the $[0, 1]$ interval, we linearly transform $SFD$ as $TSFD$ (see Equation~\ref{TSFD_eqn}; equal to $FB/FI$; $\in [0, 1]$; to be maximized)

\begin{equation} \label{TSFD_eqn}
TSFD = \frac{1+SFD}{2} = \frac{FB}{FI} 
\end{equation}

\section{Experimental Validation}
\label{sec:Experiments}

In this section, we aim to compare \textit{MaxMin Linear} to state of the art initialization methods for FCM-like clustering algorithms, i.e., \textit{MacQueen2}, Faber's, K-Means++, and repeated K-Means++ (retaining the best result). These methods are indeed the most common linear methods and are good representatives for random, probability, and distance-based methods. Moreover, they do not require any parameterization. To achieve our comparison of initialization methods, we use the indices mentioned in  Section~\ref{sec:RelatedWorks}. 

\subsection{Datasets}
\label{subsec:ExperimentDatasets}

Initialization methods are compared on 15 commonly used real-life datasets from the UCI Machine Learning Repository\footnote{\url{http://archive.ics.uci.edu/ml/}} and seven artificial datasets. Their characteristics are featured in Table~\ref{tab:table-datasets}. 

\begin{table*}
\centering
\caption{Dataset features}
\label{tab:table-datasets}
\begin{tabular}{|c|c|c|c|c|c|}
\hline
\textbf{ID} & \textbf{Datasets} & \begin{tabular}[c]{@{}c@{}} \textbf{\# of} \\ \textbf{data points} \end{tabular} & \begin{tabular}[c]{@{}c@{}}\textbf{\# of} \\ \textbf{variables} \end{tabular} & \begin{tabular}[c]{@{}c@{}} \textbf{\# of} \\ \textbf{clusters} \end{tabular} &  \textbf{Sources} \\ \hline
1 & Wine & 178 & 13  & 3 
 & UCI \\ \hline
2 & Iris & 150 & 4  & 3 
& UCI \\ \hline
3 & Seeds & 210 & 7  & 3 
 & UCI\\ \hline
4 & \begin{tabular}[c]{@{}c@{}} Original Wisconsin \\ Breast Cancer (WBCD)\end{tabular}  & 683 & 9  & 2 
 & UCI \\ \hline
5 &  \begin{tabular}[c]{@{}c@{}} Wisconsin Diagnostic \\ Breast Cancer (WDBC)\end{tabular} & 569 & 30  & 2 
 & UCI \\ \hline
6 &  BUPA Liver Disorder (BUPA)  & 345 & 6 & 2 & UCI \\ \hline
7 & Pima & 768 & 8 & 2 & UCI \\ \hline
8 & Glass & 214 & 9 & 6 & UCI \\ \hline
9 & Vehicle & 846 & 18 & 4 & UCI \\ \hline
10 & Segmentation & 2310 & 19 & 7 & UCI \\ \hline
11 & Parkinson & 150 & 22 & 2 & UCI \\ \hline
12 & Movement Libras & 360 & 90 & 15 & UCI \\ \hline
13 & Ecoli & 336 & 7 & 8 & UCI \\ \hline
14 & Yeast & 1484 & 8 & 10 & UCI \\ \hline
15 & WineQuality-Red & 1599 & 11 & 6 & UCI \\ \hline
16 & Bensaid & 49 & 2 & 3 & \cite{bensaid1996validity} \\ \hline
17 & E1071-3 & 150 & 3 & 3 & \cite{meyer2017package} \\ \hline
18 & Ruspini\_original & 75 & 2 & 4 & \cite{ruspini1970numerical} \\ \hline
19 & E1071-3-overlapped & 150 & 3 & 3 & \cite{meyer2017package} \\ \hline
20 & Ruspini\_noised & 95 & 2 & 4 & \cite{ruspini1970numerical} \\ \hline
21 & E1071-5 & 250 & 3 & 5 & \cite{meyer2017package} \\ \hline
22 & E1071-5-overlapped & 250 & 3 & 5 & \cite{meyer2017package} \\ \hline
\end{tabular}
\end{table*}

In the case of real-life datasets, the true number of clusters in each dataset is assimilated to the number of labels. Although using the number of labels as the number of clusters is debatable, it is acceptable if the set of descriptive variables explain the labels well. In artificial datasets, the number of clusters is known by construction.

In addition, we created new artificial datasets by introducing overlapping and noise to some of the existing artificial datasets such as E1071-3, Ruspini\_original, and E1071-5 datasets (see Table~\ref{tab:table-datasets}, ID 17, 18, and 21). 

To create the dataset, new data points are introduced and each must be labeled. To obtain a dataset with overlapping, we modified the construction of the E1071 artificial datasets \cite{meyer2017package}. In the original datasets, there are three or five clusters of equal size (50). Cluster $i$ is generated according to a Gaussian distribution $N(i; 0.3)$. To increase overlapping while retaining the same cluster size, we only change the standard deviation from 0.3 to 0.4. Then, there is no labeling problem.

Noise is introduced in each cluster by adding noisy points generated by a Gaussian variable around each label gravity center. First, for each label, we calculate the coordinates of centers, and the mean and standard deviation of each variable. With Gaussian variables, points mainly lie between ``center +/- two standard deviations". 

Noisy data are often generated by distributions with positive skewness. For example, in a two-dimensional dataset, for each label, we add points that are far away from the corresponding gravity center, especially on the right hand side, which generally contains the most points. Then, we draw a random number $r$ between 0 and 1. If $r \leq 0.25$, the point is attributed to the left hand side. Otherwise, the point is attributed to the right hand side. This method helps obtain noisy data that are $^1/_4$ times smaller and $^3/_4$ times greater, respectively, than the expected value for the considered label. We apply this process to the Ruspini dataset \cite{ruspini1970numerical}.

\subsection{Experimental Settings}
\label{subsec:ExperimentalSettings}

In our experiments, we parameterize the FCM algorithm as follows: default termination criterion $\epsilon = 0.0001$ and default fuzziness coefficient value $m = 2$. We used these default settings as we are only interested in improving the initialization of FCM algorithm. All initialization methods and clustering validity indices are written in Python version 2.7.4. Repeated K-Means++ runs are performed ten times. 

\subsection{Experimental Results}
\label{subsec:ExperimentalResults}

In our experiments, we compare our method \textit{MaxMin Linear} to all initialization methods from Section~\ref{sec:RelatedWorks}, on all datasets. We account for the following comparison criteria: number of iterations, $V_{PC}$, $V_{CL}$, $FB$, $FW$, $FI$, $V_{FRatio}$, $V_{TSFD}$, $V_{FS}$, and $V_{XB}$. We also rank the initialization methods with respect to all criteria.

Since presenting all results would take too much space, we only present three real-life datasets  i.e., WineQuality-Red (Tables~\ref{tab:ComparisonWineQuality}, \ref{tab:ComparisonWineQuality2}, and \ref{tab:RankingWineQuality}), Glass (Tables~\ref{tab:ComparisonGlassQuality}, \ref{tab:ComparisonGlassQuality2}, and \ref{tab:RankingGlassQuality}), and Segmentation (Tables~\ref{tab:ComparisonSegmentationQuality}, \ref{tab:ComparisonSegmentationQuality2}, and \ref{tab:RankingSegmentationQuality}), as well as two of the artificial datasets we modified to introduce noise and overlapping, i.e., Ruspini\_noised (Tables~\ref{tab:ComparisonRuspiniQuality}, \ref{tab:ComparisonRuspiniQuality2}, and \ref{tab:RankingRuspiniQuality}), and E1071-5-overlapped (Tables~\ref{tab:ComparisonMeyerQuality}, \ref{tab:ComparisonMeyerQuality2}, and \ref{tab:RankingMeyerQuality}), respectively. Finally, the average ranking of initialization methods on all datasets is presented in Table~\ref{tab:AverageRanks}.

\begin{table*}
\centering
\caption{Experiment results on WineQuality-Red (1/2)}
\label{tab:ComparisonWineQuality}
\begin{tabular}{|l|c|c|c|c|c|c|}
\hline
Initialization Method & \# of iteration & $V_{PC}$ & $V_{CL}$ & $FB$ & $FW$ \\ \hline
MacQueen2 & 45 & 0.664 & 0.7455 & 110972.7 & 1224079.7 \\ \hline
Faber & 430 & 0.664 & 0.7455 & \textbf{101440.4} & 1224079.7 \\ \hline
K-Means++ & 37 & 0.616 & 0.7029 & 101440.5 & 1089058.1 \\ \hline
K-Means++ $\times 10$ & 393 & 0.664 & 0.7455 & \textbf{101440.4} & 1224073.7 \\ \hline
\textbf{MaxMin Linear} & \textbf{34} & \textbf{0.665} & \textbf{0.7458} & 110972.7 & \textbf{1224384.8} \\ \hline
\end{tabular}
\end{table*}

\begin{table*}
\centering
\caption{Experiment results on WineQuality-Red (2/2)}
\label{tab:ComparisonWineQuality2}
\begin{tabular}{|l|c|c|c|c|c|}
\hline
Initialization Method & $FI$ & $V_{FRatio}$ & $V_{TSFD}$ & $V_{FS}$ & $V_{XB}$ \\ \hline
MacQueen2 & 1335052.363 & 11.0305 & \textbf{0.9169} & -1113107.01 & 0.1621 \\ \hline
Faber & 1335052.363 & 11.0305 & 0.9148 & -1113107.01 & 0.1621 \\ \hline
K-Means++ & 1190498.537 & 10.7359 & 0.9148 & -987617.57 & 0.2388 \\ \hline
K-Means++ $\times 10$ & 1335046.425 & 11.0304 & 0.9148 & -1113101.04 & 0.1621 \\ \hline
\textbf{MaxMin Linear} & \textbf{1335357.554} & \textbf{11.0332} & \textbf{0.9169} & \textbf{-1113412.13} & \textbf{0.1611} \\ \hline
\end{tabular}
\end{table*}

\begin{table*}
\centering
\caption{Ranking of initialization methods on WineQuality-Red}
\label{tab:RankingWineQuality}
\begin{tabular}{|l|c|c|c|c|c|c|c|c|c|c|}
\hline
Initialization Method & \begin{tabular}[c]{@{}c@{}}\# of \\ iteration \end{tabular} & $V_{PC}$ & $V_{CL}$ & $FB$ & $FW$ & $FI$ & $V_{FRatio}$ & $V_{TSFD}$ & $V_{FS}$ & $V_{XB}$ \\ \hline
MacQueen2  & 3 & 2 & 2 & 4 & 2 & 2 & 2 & 2 & 2 & 2 \\ \hline
Faber  & 5 & 2 & 2 & 2 & 2 & 2 & 2 & 5 & 2 & 2 \\ \hline
K-Means++  & 2 & 5 & 5 & 3 & 5 & 5 & 5 & 3 & 5 & 5 \\ \hline
K-Means++ $\times 10$ & 4 & 4 & 4 & \textbf{1} & 4 & 4 & 4 & 4 & 4 & 4 \\ \hline
\textbf{MaxMin Linear}  & \textbf{1} & \textbf{1} & \textbf{1} & 5 & \textbf{1} & \textbf{1} & \textbf{1} & \textbf{1}  & \textbf{1} & \textbf{1} \\ \hline
\end{tabular}
\end{table*}

\begin{table*}
\centering
\caption{Experiment results on Glass (1/2)}
\label{tab:ComparisonGlassQuality}
\begin{tabular}{|l|c|c|c|c|c|}
\hline
Initialization Method & \begin{tabular}[c]{@{}c@{}}\# of \\ iteration \end{tabular} & $V_{PC}$ & $V_{CL}$ & $FB$ & $FW$  \\ \hline
MacQueen2 & \textbf{44} & 0.493 & 0.570 & 452.6 & \textbf{154.1} \\ \hline
Faber & 456 & 0.493 & 0.570 & 452.6 & \textbf{154.1} \\ \hline
Kmeans++ & 56 & 0.493 & 0.570 & 452.6 & \textbf{154.1} \\ \hline
K-Means++ $\times 10$ & 366 & 0.493 & 0.570 & 452.6 & \textbf{154.1} \\ \hline
\textbf{MaxMin Linear} & 68 & \textbf{0.555} & \textbf{0.645} & \textbf{508.3} & 162.9 \\ \hline
\end{tabular}
\end{table*}

\begin{table*}
\centering
\caption{Experiment results on Glass (2/2)}
\label{tab:ComparisonGlassQuality2}
\begin{tabular}{|l|c|c|c|c|c|}
\hline
Initialization Method & $FI$ & $V_{FRatio}$ & $V_{TSFD}$ & $V_{FS}$ & $V_{XB}$ \\ \hline
MacQueen2 & 606.8 & 2.94 & 0.74596 & -298.5 & 2.358 \\ \hline
Faber & 606.8 & 2.94 & 0.74597 & -298.5 & 2.358 \\ \hline
Kmeans++ & 606.7 & 2.94 & 0.74593 & -298.4 & 2.358 \\ \hline
K-Means++ $\times 10$ & 606.7 & 2.94 & 0.74604 & -298.4 & 2.358 \\ \hline
\textbf{MaxMin Linear} & \textbf{671.2} & \textbf{3.12} & \textbf{0.75725} & \textbf{-345.4} & \textbf{0.453} \\ \hline
\end{tabular}
\end{table*}

\begin{table*}
\centering
\caption{Ranking of initialization methods on Glass}
\label{tab:RankingGlassQuality}
\begin{tabular}{|l|c|c|c|c|c|c|c|c|c|c|}
\hline
Initialization method & \begin{tabular}[c]{@{}c@{}}\# of \\ iteration \end{tabular} & $V_{PC}$ & $V_{CL}$ & $FB$ & $FW$ & $FI$ & $V_{FRatio}$ & $V_{TSFD}$ & $V_{FS}$ & $V_{XB}$ \\ \hline
MacQueen2 & \textbf{1} & 2 & 2 & 2 & \textbf{1} & 2 & 2 & 4 & 2 & 5 \\ \hline
Faber & 5 & 3 & 3 & 3 & 2 & 3 & 3 & 3 & 3 & 2 \\ \hline
Kmeans++ & 2 & 5 & 5 & 5 & 4 & 5 & 5 & 5 & 5 & 4 \\ \hline
K-Means++ $\times 10$  & 4 & 4 & 4 & 4 & 3 & 4 & 4 & 2 & 4 & 3 \\ \hline
\textbf{MaxMin Linear} & 3 & \textbf{1} & \textbf{1} & \textbf{1} & 5 & \textbf{1} & \textbf{1} & \textbf{1} & \textbf{1} & \textbf{1} \\ \hline
\end{tabular}
\end{table*}

\begin{table*}
\centering
\caption{Experiment results on Segmentation (1/2)}
\label{tab:ComparisonSegmentationQuality}
\begin{tabular}{|l|c|c|c|c|c|}
\hline
Initialization Method & \begin{tabular}[c]{@{}c@{}}\# of \\ iteration \end{tabular} & $V_{PC}$ & $V_{CL}$ & $FB$ & $FW$  \\ \hline
MacQueen2 & 103 & 0.381 & 0.476 & 12384361.4 & 5781042.6 \\ \hline
Faber & 731 & 0.398 & 0.488 & 14157566.6 & 5680259.6 \\ \hline
Kmeans++ & 146 & 0.381 & 0.476 & 12388277.9 & 5781061.6 \\ \hline
K-Means++ $\times 10$ & 930 & 0.399 & 0.490 & 14254025.9 & \textbf{5666840.5} \\ \hline
\textbf{MaxMin Linear} & \textbf{54} & \textbf{0.430} & \textbf{0.526} & \textbf{19234921.0} & 6344612.7 \\ \hline
\end{tabular}
\end{table*}

\begin{table*}
\centering
\caption{Experiment results on Segmentation (2/2)}
\label{tab:ComparisonSegmentationQuality2}
\begin{tabular}{|l|c|c|c|c|c|}
\hline
Initialization Method & $FI$ & $V_{FRatio}$ & $V_{TSFD}$ & $V_{FS}$ & $V_{XB}$ \\ \hline
MacQueen2 & 18165404.0 & 2.14 & 0.6818 & -6603318.7 & 0.363 \\ \hline
Faber & 19837826.2 & 2.49 & 0.7137 & -8477307.1 & 0.464 \\ \hline
Kmeans++ & 18169339.6 & 2.14 & 0.6818 & -6607216.3 & 0.361 \\ \hline
K-Means++ $\times 10$ & 19920866.4 & 2.52 & 0.7136 & -8587185.5 & \textbf{0.341} \\ \hline
\textbf{MaxMin Linear} & \textbf{25579533.7} & \textbf{3.03} & \textbf{0.7520} & \textbf{-12890308.3} & 0.656 \\ \hline
\end{tabular}
\end{table*}

\begin{table*}
\centering
\caption{Ranking of initialization methods on Segmentation}
\label{tab:RankingSegmentationQuality}
\begin{tabular}{|l|c|c|c|c|c|c|c|c|c|c|}
\hline
Initialization method & \begin{tabular}[c]{@{}c@{}}\# of \\ iteration \end{tabular} & $V_{PC}$ & $V_{CL}$ & $FB$ & $FW$ & $FI$ & $V_{FRatio}$ & $V_{TSFD}$ & $V_{FS}$ & $V_{XB}$ \\ \hline
MacQueen2 & 2 & 4 & 5 & 5 & 3 & 5 & 5 & 5 & 5 & 3 \\ \hline
Faber & 4 & 3 & 3 & 3 & 2 & 3 & 3 & 2 & 3 & 4 \\ \hline
Kmeans++ & 3 & 5 & 4 & 4 & 4 & 4 & 4 & 3 & 4 & 2 \\ \hline
K-Means++ $\times 10$  & 5 & 2 & 2 & 2 & \textbf{1} & 2 & 2 & 4 & 2 & \textbf{1} \\ \hline
\textbf{MaxMin Linear} & \textbf{1} & \textbf{1} & \textbf{1} & \textbf{1} & 5 & \textbf{1} & \textbf{1} & \textbf{1} & \textbf{1} & 5 \\ \hline
\end{tabular}
\end{table*}

\begin{table*}
\centering
\caption{Experiment results on Ruspini\_noised (1/2)}
\label{tab:ComparisonRuspiniQuality}
\begin{tabular}{|l|c|c|c|c|c|}
\hline
Initialization Method  & \begin{tabular}[c]{@{}c@{}}\# of \\ iteration \end{tabular} & $V_{PC}$ & $V_{CL}$ & $FB$ & $FW$ \\ \hline
MacQueen2 & 9 & 0.775121 & 0.806518 & 219099.6 & 23421.0260  \\ \hline
Faber & 130 & 0.775125 & 0.806517 & 219100.8 & 23421.0258  \\ \hline
Kmeans++ & 13 & 0.775122 & 0.806521 & 219101.1 & 23421.0258 \\ \hline
K-Means++ $\times 10$  & 105 & \textbf{0.775128} & 0.806518 & 219102.3 & \textbf{23421.0256}  \\ \hline
\textbf{MaxMin Linear} & \textbf{7} & 0.775128 & \textbf{0.806523} & \textbf{219105.4} & 23421.0268 \\ \hline
\end{tabular}
\end{table*}

\begin{table*}
\centering
\caption{Experiment results on Ruspini\_noised (2/2)}
\label{tab:ComparisonRuspiniQuality2}
\begin{tabular}{|l|c|c|c|c|c|}
\hline
Initialization Method  & $FI$ & $V_{FRatio}$ & $V_{TSFD}$ & $V_{FS}$ & $V_{XB}$ \\ \hline
MacQueen2 & 242520.7 & 9.3548 & 0.903427 & -195678.6 & 0.063680 \\ \hline
Faber & 242521.9 & 9.3549 & 0.903427 & -195679.8 & 0.063681 \\ \hline
Kmeans++ & 242522.1 & 9.3549 & 0.903427 & -195680.0 & 0.063676 \\ \hline
K-Means++ $\times 10$ & 242523.3 & 9.3549 & 0.903426 & -195681.3 & 0.063681 \\ \hline
\textbf{MaxMin Linear} & \textbf{242526.4} & \textbf{9.3551} & \textbf{0.903429} & \textbf{-195684.4} & \textbf{0.063672} \\ \hline
\end{tabular}
\end{table*}

\begin{table*}
\centering
\caption{Ranking of initialization methods on Ruspini\_noised}
\label{tab:RankingRuspiniQuality}
\begin{tabular}{|l|c|c|c|c|c|c|c|c|c|c|}
\hline
Initialization Method & \begin{tabular}[c]{@{}c@{}}\# of \\ iteration \end{tabular} & $V_{PC}$ & $V_{CL}$ & $FB$ & $FW$ & $FI$ & $V_{FRatio}$ & $V_{TSFD}$ & $V_{FS}$ & $V_{XB}$ \\ \hline
MacQueen2 & 2 & 5 & 3 & 5 & 4 & 5 & 5 & 4 & 5 & 3 \\ \hline
Faber & 5 & 3 & 5 & 4 & 2 & 4 & 4 & 3 & 4 & 5 \\ \hline
Kmeans++ & 3 & 4 & 2 & 3 & 3 & 3 & 3 & 2 & 3 & 2 \\ \hline
K-Means++ $\times 10$ & 4 & \textbf{1} & 4 & 2 & \textbf{1} & 2 & 2 & 5 & 2 & 4 \\ \hline
\textbf{MaxMin Linear} & \textbf{1} & 2 & \textbf{1} & \textbf{1} & 5 & \textbf{1} & \textbf{1} & \textbf{1} & \textbf{1} & \textbf{1} \\ \hline
\end{tabular}
\end{table*}

\begin{table*}
\centering
\caption{Experiment results on E1071-5-overlapped (1/2)}
\label{tab:ComparisonMeyerQuality}
\begin{tabular}{|l|c|c|c|c|c|}
\hline
Initialization Method & \begin{tabular}[c]{@{}c@{}}\# of \\ iteration \end{tabular} & $V_{PC}$ & $V_{CL}$ & $FB$ & $FW$  \\ \hline
MacQueen2 & 8 & 0.735646 & 0.762681 & 219.7337 & 48.715631 \\ \hline
Faber & 103 & 0.735645 & 0.762683 & 219.7358 & 48.715630 \\ \hline
Kmeans++ & 12 & 0.735651 & 0.762685 & 219.7408 & 48.715632 \\ \hline
K-Means++ $\times 10$ & 113 & 0.735645 & 0.762683 & 219.7363 & \textbf{48.715629} \\ \hline
\textbf{MaxMin Linear} & \textbf{7} & \textbf{0.735652} & \textbf{0.762688} & \textbf{219.7445} & \textbf{48.715629} \\ \hline
\end{tabular}
\end{table*}

\begin{table*}
\centering
\caption{Experiment results on E1071-5-overlapped (2/2)}
\label{tab:ComparisonMeyerQuality2}
\begin{tabular}{|l|c|c|c|c|c|}
\hline
Initialization Method & $FI$ & $V_{FRatio}$ & $V_{TSFD}$ & $V_{FS}$ & $V_{XB}$ \\ \hline
MacQueen2 & 268.4494 & 4.5105 & 0.818530 & -171.0181 & 0.11574 \\ \hline
Faber & 268.4514 & 4.5106 & 0.818535 & -171.0202 & \textbf{0.11569} \\ \hline
Kmeans++ & 268.4565 & 4.5107 & 0.818534 & -171.0252 & 0.11575 \\ \hline
K-Means++ $\times 10$ & 268.4519 & 4.5106 & 0.818530 & -171.0207 & \textbf{0.11569} \\ \hline
\textbf{MaxMin Linear} & \textbf{268.4601} & \textbf{4.5108} & \textbf{0.818537} & \textbf{-171.0288} & 0.11572 \\ \hline
\end{tabular}
\end{table*}

\begin{table*}
\centering
\caption{Ranking of initialization methods on E1071-5-overlapped}
\label{tab:RankingMeyerQuality}
\begin{tabular}{|l|c|c|c|c|c|c|c|c|c|c|}
\hline
Initialization method & \begin{tabular}[c]{@{}c@{}}\# of \\ iteration \end{tabular} & $V_{PC}$ & $V_{CL}$ & $FB$ & $FW$ & $FI$ & $V_{FRatio}$ & $V_{TSFD}$ & $V_{FS}$ & $V_{XB}$ \\ \hline
MacQueen2 & 2 & 3 & 5 & 5 & 4 & 5 & 5 & 5 & 5 & 4 \\ \hline
Faber & 4 & 5 & 4 & 4 & 3 & 4 & 4 & 2 & 4 & \textbf{1} \\ \hline
Kmeans++ & 3 & 2 & 2 & 2 & 5 & 2 & 2 & 3 & 2 & 5 \\ \hline
K-Means++ $\times 10$  & 5 & 4 & 3 & 3 & \textbf{1} & 3 & 3 & 4 & 3 & 2 \\ \hline
\textbf{MaxMin Linear} & \textbf{1} & \textbf{1} & \textbf{1} & \textbf{1} & 2 & \textbf{1} & \textbf{1} & \textbf{1} & \textbf{1} & 3 \\ \hline
\end{tabular}
\end{table*}

\begin{table*}
\centering
\caption{Average ranking of initialization methods on all datasets}
\label{tab:AverageRanks}
\begin{tabular}{|l|c|c|c|c|c|c|c|c|c|c|}
\hline
Initialization method & \begin{tabular}[c]{@{}c@{}}\# of \\ iteration \end{tabular} & $V_{PC}$ & $V_{CL}$ & $FB$ & $FW$ & $FI$ & $V_{FRatio}$ & $V_{TSFD}$ & $V_{FS}$ & $V_{XB}$ \\ \hline
MacQueen2  & 1.95 & 3.36 & 3.55 & 3.86 & 3.41 & 3.41 & 3.41 & 3.04 & 3.41 & 3.55 \\ \hline
Faber& 4.45 & 2.73 & 2.82 & 1.73 & 2.73& 2.73 & 2.73 & 3.27 & 2.73 & 2.91 \\ \hline
K-Means++  &  1.95 & 3.86 & 3.68 & 3.86 & 3.86  & 3.86 & 3.86  & 3.54 & 3.86 & 3.36 \\ \hline
K-Means++ $\times 10$ &  4.41 & 2.68 & 2.55 & \textbf{1.64} & 2.86  & 2.86 & 2.86 & 3.22 & 2.86 & 2.82 \\ \hline
\textbf{MaxMin Linear} & \textbf{1.68} & \textbf{2.27} & \textbf{2.32} & 3.82 & \textbf{2.05}  & \textbf{2.05} & \textbf{2.05} & \textbf{1.86} & \textbf{2.05} & \textbf{2.27} \\ \hline
\end{tabular}
\end{table*}



From these experimental results, several observations can be drawn. In regard to the number of iterations, recall that Faber's and K-Means++ $\times 10$ methods are relaunches of two stochastic initialization methods: \textit{MacQueen2} and K-Means++, respectively. With an average ranking of 1.68 (Table~\ref{tab:AverageRanks}),  \textit{MaxMin Linear} outperforms all other methods, including single-run methods \textit{MacQueen2} (average ranking: 1.95) and K-Means++ (average ranking: 1.95).

Regarding clustering result quality, \textit{MaxMin Linear} obtains the best average ranking for eight of the nine experimented quality indices  (Table~\ref{tab:AverageRanks}). Only the $FB$ index yields a better result for the two multiple-runs methods, while the result of \textit{Maxmin Linear} is similar to those of \textit{MacQueen2} and K-Means++. However, \textit{Maxmin Linear} achieves the best trade-off between $FB$ and $FW$, and thus maximizes the indices that take  both $FB$ and $FW$ into account ($V_{FRatio}$, $V_{TSFD}$, $V_{FS}$ and $V_{XB}$). The best result for \textit{MaxMin Linear} is obtained with $V_{TSFD}$ (average ranking of 1.86; Table~\ref{tab:AverageRanks}), the new index specially tailored for fuzzy clustering that we propose.

In conclusion, the results obtained with \textit{MaxMin Linear} are a little better than those obtained with multiple-runs methods, but they require ten times fewer iterations. Moreover, \textit{MaxMin Linear} is deterministic, whereas  multiple-runs methods are stochastic.

\section{Conclusion and Perspectives}
\label{sec:Conclusion}

In this paper, we propose a new, fast, and easy to implement initialization method for FCM called \textit{MaxMin Linear}. \textit{MaxMin Linear} is compared to several initialization methods from the literature. It is experimentally shown that \textit{MaxMin Linear} outperforms existing methods on 22 datasets. Moreover, we also propose an appropriate fuzzy validity index, TSFD, to evaluate initialization methods.

In addition, \textit{MaxMin Linear} can be applied to algorithms other than FCM, such as Fuzzy K-Modes and Fuzzy K-Medoids, which apply on categorical. In particular, \textit{MaxMin Linear} allows decreasing the complexity of Park's Fuzzy K-Medoids implementation. 

In consequence, an immediate perspective to our work is to propose a new clustering ensemble method for heterogeneous datasets composed of both numerical and categorical data. 

\section*{Acknowledgments} This project is supported by the Rh\^{o}ne Alpes Region's ARC~5: ``Cultures, Sciences, Soci\'et\'es et M\'ediations'' through A. \"Ozt\"urk's Ph.D. grant.

\bibliographystyle{splncs}
\bibliography{references}

\end{document}